%% file: main.tex
\documentclass[10pt,twocolumn,letterpaper]{article}

\usepackage{iccv}              


\usepackage{stfloats}
\usepackage{float}

\definecolor{iccvblue}{rgb}{0.21,0.49,0.74}
\usepackage[pagebackref,breaklinks,colorlinks,allcolors=iccvblue]{hyperref}

\def\paperID{22} 
\def\confName{ICCV}
\def\confYear{2025}

\title{UNet-Based Keypoint Regression for 3D Cone Localization\\ in Autonomous Racing}

\author{
Mariia Baidachna$^{1*}$ \quad
James Carty$^{2}$ \quad
Aidan Ferguson$^{1}$ \quad
Joseph Agrane$^{1}$ \quad
Varad Kulkarni$^{1}$ \quad \\
Aubrey Agub$^{1}$ \quad
Michael Baxendale$^{1}$ \quad
Aaron David$^{1}$ \quad
Rachel Horton$^{1}$ \quad
Elliott Atkinson$^{1}$ \\
$^{1}$School of Computer Science, University of Glasgow, G12 8QQ, UK \\
$^{2}$Amazon, Edinburgh, UK \\
{\tt\small 2828197b@student.gla.ac.uk}
}

\begin{document}
\maketitle
\input{sec/0_abstract}    
\input{sec/1_intro}

\input{sec/2_methods}
\input{sec/3_results}
\input{sec/4_conclusion}
{
    \small
    \bibliographystyle{ieeenat_fullname}
    \bibliography{main}
}

\end{document}

%% file: sec/0_abstract.tex
\begin{abstract}

Accurate cone localization in 3D space is essential in autonomous racing for precise navigation around the track. Approaches that rely on traditional computer vision algorithms are sensitive to environmental variations, and neural networks are often trained on limited data and are infeasible to run in real time. We present a UNet-based neural network for keypoint detection on cones, leveraging the largest custom-labeled dataset we have assembled. Our approach enables accurate cone position estimation and the potential for color prediction. Our model achieves substantial improvements in keypoint accuracy over conventional methods. Furthermore, we leverage our predicted keypoints in the perception pipeline and evaluate the end-to-end autonomous system. Our results show high-quality performance across all metrics, highlighting the effectiveness of this approach and its potential for adoption in competitive autonomous racing systems.

\end{abstract}

%% file: sec/1_intro.tex
\section{Introduction}
\label{sec:intro}
Accurate cone detection in 3D space is critical for success in autonomous racing. Particularly, the competitive environment of racing competitions, such as Formula Student, where teams worldwide participate with autonomous racing systems, demands fast and precise cone localization in order for the car to navigate a previously unseen track. The boundaries of the track are defined by blue cones on the left and yellow cones on the right, making cone position and color identification a pivotal point in the autonomous racing system. For this task, Keypoint Regression (KPR) techniques are pivotal. They excel at fine-grained detection of cone features \cite{Sun2018}, enabling localization by maximizing the information extracted from data, ultimately leading to more reliable and safer decision-making on the racetrack.

Computer vision methods are integral to solving the task of autonomous racing, allowing autonomous vehicles (AVs) to perceive complex environments in real-time. Object detection is a key component of this, which ensures safe and efficient navigation across dynamic settings. In particular, KPR techniques, which often make use of convolutional neural networks (CNNs) \cite{Sun2018, Zhang2019} and more recently transformer architectures \cite{Liang2023, Yu2024}, have become prevalent in object detection across a wide array of applications. 

Real-time cone perception is complicated by several unique challenges. Cones are small with varying distances from the vehicle, making them difficult to detect accurately, especially at high speeds and in the wild \cite{s24227347}. Their condition and appearance can also vary widely. This can be due to collision damage, physical marks such as stains and scuffs, or environmental exposure like mud and weathering. Varying environmental and light conditions add further complexity to the cone detection task \cite{mohammed2020perception, fursa2021worsening}. These challenges, therefore, make the development of robust perception algorithms crucial for maintaining the safety and performance of AVs while racing.

To overcome the obstacles of KPR and position estimation within autonomous racing, we propose our contributions:

\begin{itemize}
    \item \textbf{Dataset}: The largest publicly available labeled dataset of 25k annotated cone images, captured from various perspectives and conditions, published at \href{https://www.kaggle.com/datasets/mashapotatoes/cone-dataset}{Kaggle}\footnote{Dataset available at: \url{https://www.kaggle.com/datasets/mashapotatoes/cone-dataset}}.

    \item \textbf{Novel KPR Method}: A novel UNet-based architecture for KPR, specifically designed to accurately localize keypoints in cones within complex scenes.
    
    \item \textbf{Application in Autonomous Navigation}: We integrate our KPR model within an autonomous vehicle perception system, providing methods on how to use it to enhance the overall performance and evaluating it based on simulations.
\end{itemize}

Importantly, these additions to our pipeline allow our system to leverage geometric constraints (via stereo disparity) to derive 3D supervision without requiring ground-truth 3D annotations, making it scalable to larger datasets. While prior methods like NeRF \cite{mildenhall2020nerf} and monocular depth estimation \cite{eigen2014depth} aim to reconstruct dense 3D scenes, our approach offers a lightweight and accurate alternative tailored for targeted localization of key 3D landmarks in dynamic environments within the autonomous racing context.

\subsection{Related Work}
Despite the critical role of cones in certain AV environments, most state-of-the-art methods for object detection for AVs, such as those covered in the survey from Contreras et al. \cite{Contreras2024}, focus on pedestrians, cars, and cyclists, with relatively little emphasis on the detection of cones. Although this gap remains largely unaddressed in autonomous racing, prior research has explored cone detection in specific scenarios. For instance, Katsamenis et al. \cite{Katsamenis2022} applied the YOLOv5 object detection to real-time cone detection using roadwork images, achieving high accuracy with an IoU score of up to 91.31\%. 

Other approaches have also been applied for autonomous racing applications, making use of monocular cameras and operating on the Jetson TX2. YOLOv2 used by Ankit Dhall et al. {\cite{Dhall2019}} makes use of a Residual Neural Network (ResNet) to carry out the KPR. More recently, SNET-3L, a UNet-based architecture developed by Albaranez-Martinez et al. \cite{Martinez2022}, has been used to predict bounding boxes around cones.

In broader applications, a KPR strategy was used to modify YOLOv4 \cite{Wang2023}, enhancing bounding box regression and improving detection of small objects. This modification resulted in an average precision score of 45.6\% on the COCO2017, compared to 45.4\% for YOLOv4. The authors in \cite{maji2022yolo} enhance YOLO for multi-person position estimation, adding the Object Keypoint Similarity (OKS) algorithm to the loss function throughout training. They achieve state-of-the-art performance on COCO \cite{lin2014microsoft}.

KPR-based models which make use of keypoint heatmaps have also been investigated in the literature, such as CenterNet \cite{Duan2024} , which use them to find object center points and by OSKDet \cite{Lu2022} to extract information about object orientation. Prior to the development of deep learning methods, algorithmic approaches were utilized for keypoint detection and description. Notably, the Scale Invariant Feature Transform (SIFT) \cite{lowe2004sift} strengthened robustness to scale and rotation  \cite{Nguyen2014, Nandhini2023}, and Features from Accelerated Segment Test (FAST) improved computational efficiency enabling real-time use cases. \cite{Rosten2006, Rosten2010}. 

\subsubsection{Current Limitations} 

Although many KPR methods rely on traditional feature-matching methods, such as SIFT \cite{Nguyen2014, Nandhini2023} and SURF (Speeded Up Robust Features) \cite{bay2008speeded}, that have long been used for object detection in computer vision, deep learning models often outperform them \cite{bellavia2019evaluation, o2020deep}. These methods face significant limitations when applied to autonomous racing environments, where conditions are dynamic and real-time processing is essential. This includes damage, stains, or obscuring marks that make reliable detection more challenging.

Recent evaluations of image-matching algorithms, such as those by Fursa et al. \cite{fursa2021worsening}, are consistent with these claims, confirming that traditional feature-matching algorithms like SIFT and SURF are less reliable in autonomous racing scenarios, where objects are small and exposed to high variability in appearance due to changing light and weather conditions. Although newer algorithms such as SuperGlue \cite{sarlin2020superglue} leverage graph neural networks to improve feature matching, they still underperform in identifying objects with unique, application-specific shapes like cones, especially without fine-tuning on cone-specific features. While SuperGlue represents a significant advancement, its general-purpose design can limit accuracy and efficiency in detecting cones, which require highly targeted features for robust detection at high speeds. Furthermore, these methods often require offline inference, making them unsuitable for on-device AV perception tasks \cite{sarlin2020superglue}.

In contrast, deep learning models excel at capturing highly discriminative, application-specific features while maintaining robustness to high variability in complex environments \cite{he2016deep}. Considering these advantages, we have chosen to pursue a deep learning approach in this work, as it aligns with the demands of accurate cone detection in autonomous racing.

%% file: sec/2_methods.tex
\section{Methods}
\label{sec:methods}

\subsection{System Background}
\begin{figure*}[ht]
    \centering
    \includegraphics[width=0.8\linewidth]{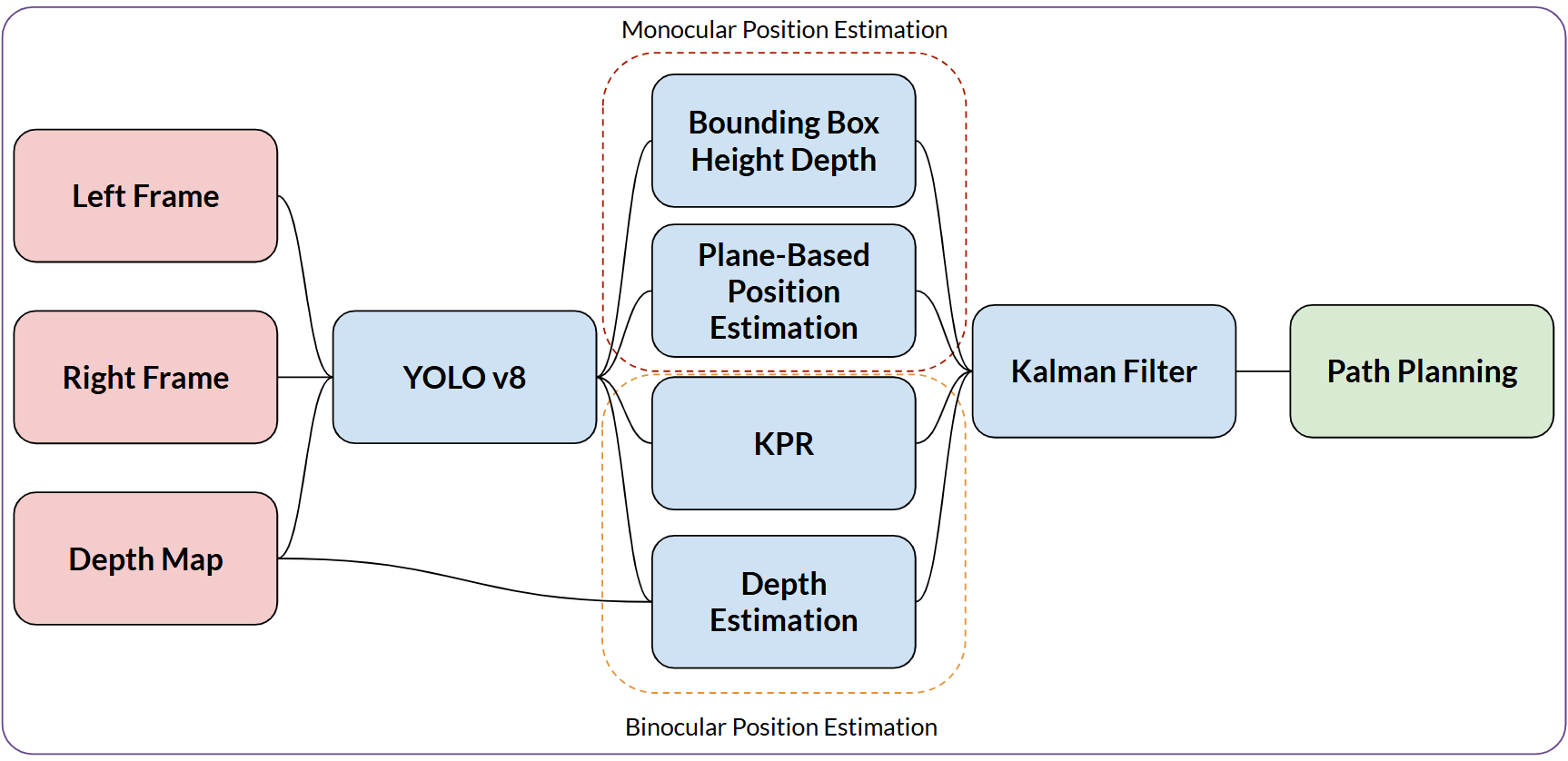}
    \caption{ Overview of the camera pipeline. Different cone position estimation methods are performed in parallel and combined with an Extended Kalman Filter. Red and Green represent system inputs and outputs respectively, with blue representing pipeline sub-routines.}
    \label{camera_pipeline}
\end{figure*}

\begin{figure}
    \centering
    \includegraphics[width=1\linewidth]{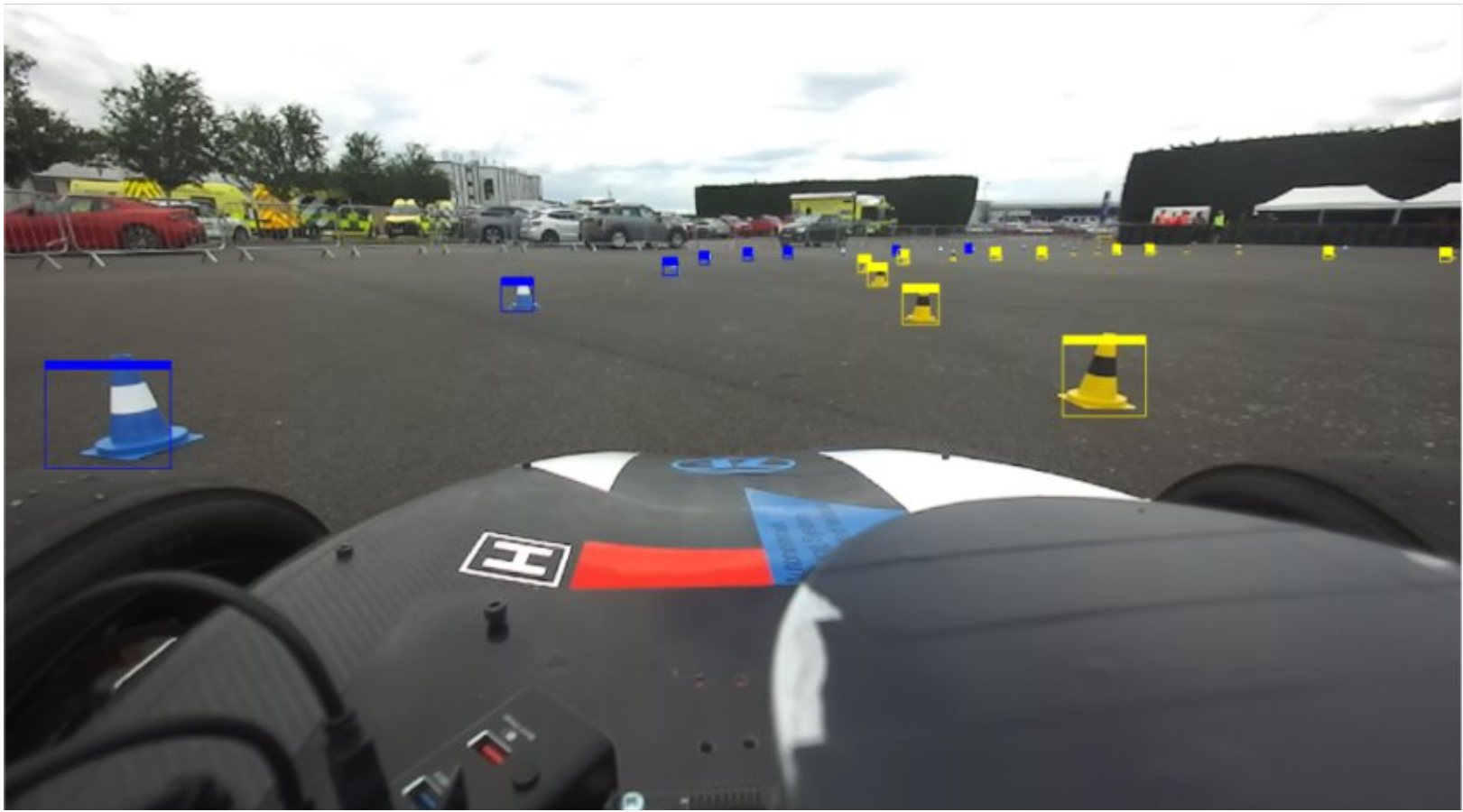}
    \caption{ A ZED2 left camera frame with YOLOv8 bounding boxes. Solid bars above cone annotations represent the confidence of the YOLOv8 predictions.}
    \label{camera_view}
\end{figure}

Since KPR is part of a larger system, we provide the necessary system overview for context on how it works. The Stereolab's ZED2 camera frames are processed through YOLOv8, the entry point in the perception pipeline, as depicted in Figure \ref{camera_pipeline}. In this paper, we focus on developing the KPR method with neural networks and justifying its addition to the perception pipeline. The initial detection step from the perspective of the ZED2 camera is visualized in Figure \ref{camera_view}, where YOLOv8 identifies and highlights cones with bounding boxes in the left camera frame. 

The bounding box information from YOLOv8 is then passed on to a multitude of position estimation methods, which estimate cone positions. Experimentally derived covariances are used to combine each estimate. The robustness of this perception stage is essential, as it feeds directly into path planning, where reliable cone detection and position estimation are critical. Each method in this system contributes uniquely, enhancing the pipeline's reliability under unpredictable conditions and sensor failure.

\subsection{Dataset Curation}

\begin{figure}
    \centering
    \includegraphics[width=1\linewidth]{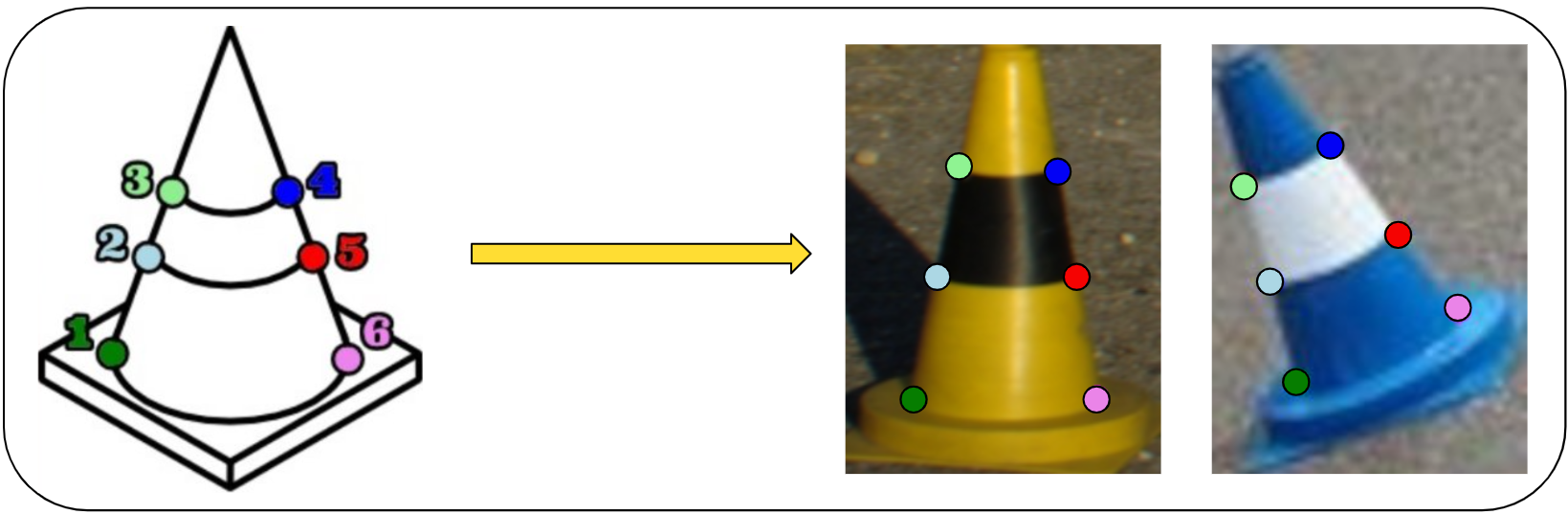}
    \caption{ General keypoint labels and their corresponding labels on a blue and yellow cone. }
    \label{cone_labels}
\end{figure}

Our dataset was developed using a custom annotation tool built with Flask. It aids the process of labeling images of cones for keypoint detection and integrates with AWS S3 buckets. The application initiates by populating lists of available images and the annotated images saved as JSON files. Each annotation includes six keypoints per cone image, as shown in Figure \ref{cone_labels}, for a total of 24,904. Upon filtering out improperly labeled images, such as an incorrect number of points or imprecise labels, we obtain a high-quality subset of 20k samples. The labeled images are of diverse quality and color, and are representative of the bounding boxes detected with YOLOv8, based on the FSOCO \cite{fsoco_2022} dataset. 

While a minimum of one point would suffice for basic position estimation and four keypoints for basic estimation of cone color classification, the additional keypoints allow for robustness to keypoint inaccuracies and the option to cross-validate. For flexibility, the model can also be trained on just the four boundary keypoints surrounding the cone strip. In any case, the keypoints are defined by the points around the strip and the base of the cone and are typically easier for the network to locate than more homogeneous regions.

\subsection{Model Architecture and Training}

\begin{figure}
    \centering
    \includegraphics[width=0.8\linewidth]{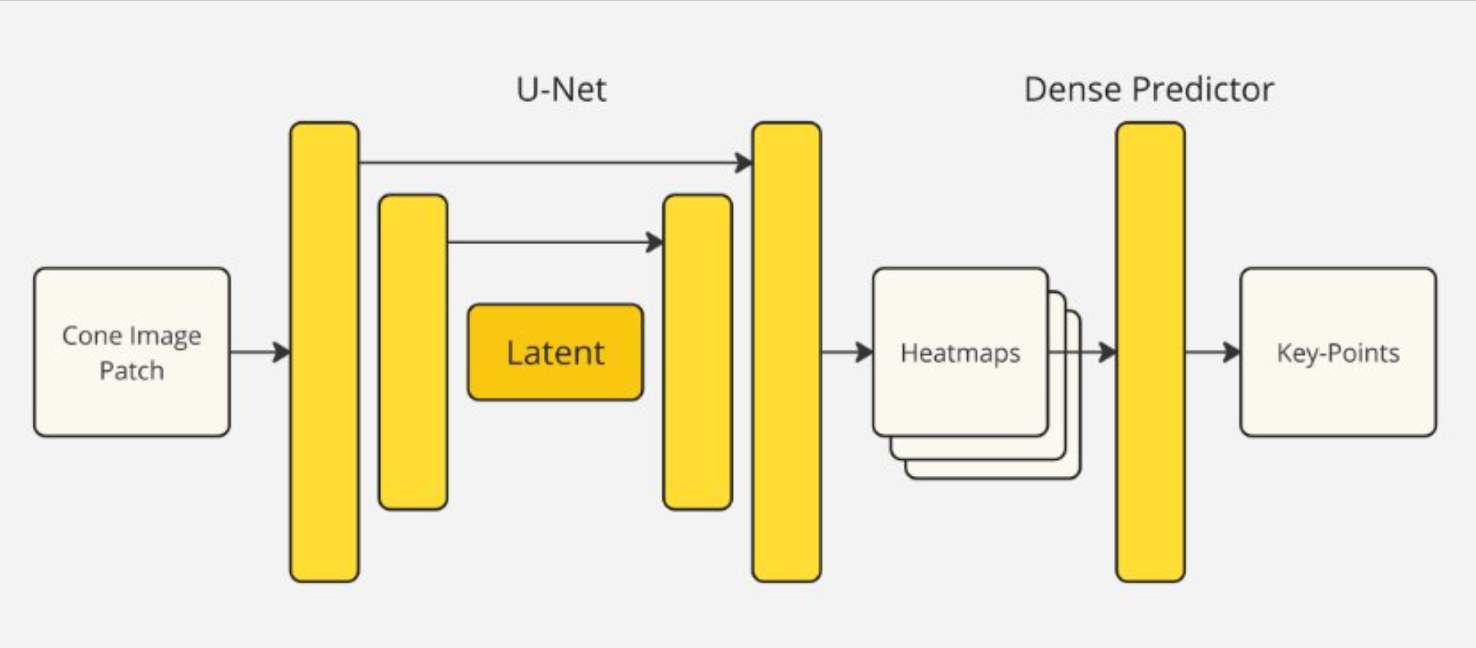}
    \caption{ A high-level overview of the architecture used in our KPR model. }
    \label{architecture_pipeline}
\end{figure}

Our proposed model utilizes a UNet architecture, composed of encoder, decoder, and bottleneck layers, as shown in Figure \ref{architecture_pipeline}. Each block in the UNet is a combination of two 2D convolutional layers with batch normalization and ReLU activation. The encoder progressively reduces spatial dimensions through downsampling, while the decoder increases them to reconstruct the output. The dimensions, strides, and padding for each convolutional block are determined by the input resolution and follow a consistent pattern. Specifically, the convolutional layers use 3x3 kernels with a stride of 2 for downsampling and a stride of 1 for maintaining resolution in the decoder. Padding is set to 1 to ensure the output size matches the expected dimensions. The number of output channels doubles as we move down the encoder (e.g., 64 $\rightarrow$ 128 $\rightarrow$ 256 $\rightarrow$ 512) and halves during upsampling in the decoder. The final layer outputs are normalized and applied to a linear prediction layer. The forward pass in the model sequentially applies the encoder and decoder layers to output the predicted keypoints.

To improve model robustness and accuracy, we perform data augmentation via rotations and randomized boundary cropping, which helps reduce overfitting and improves generalization. Four transformations were used: \texttt{NONE}, \texttt{ROTATE\_90}, \texttt{ROTATE\_180}, and \texttt{ROTATE\_270}. These rotations adjust the data in both image and keypoint coordinates. Using rotation matrices, we map keypoints to their respective locations after transformation, ensuring accurate position data for each augmented image.


To build a dataset suitable for model training, we preprocess images and annotations for the model. We first filter out any images lacking sufficient keypoints or those marked as rejected in the metadata. Selected images are further augmented with transformations and assigned unique IDs. The dataset is split into 70\% training, 20\% validation, and 10\% testing subsets, allowing performance evaluation on unseen data.


For model training, we use the \texttt{UNet} network with the configuration parameters of three channels, six keypoints, 80$\times$80 images, and a dropout rate of 0.3. The model is optimized using the AdamW \cite{loshchilov2019decoupledweightdecayregularization} optimizer with an exponential learning rate scheduler, which reduces the learning rate by a factor of \(0.99\) per epoch. The loss function combines heatmap-based and position-based losses, with \texttt{CustomLoss} supporting both L1 and smooth L1 losses.

\begin{figure}
    \centering
    \includegraphics[width=1\linewidth]{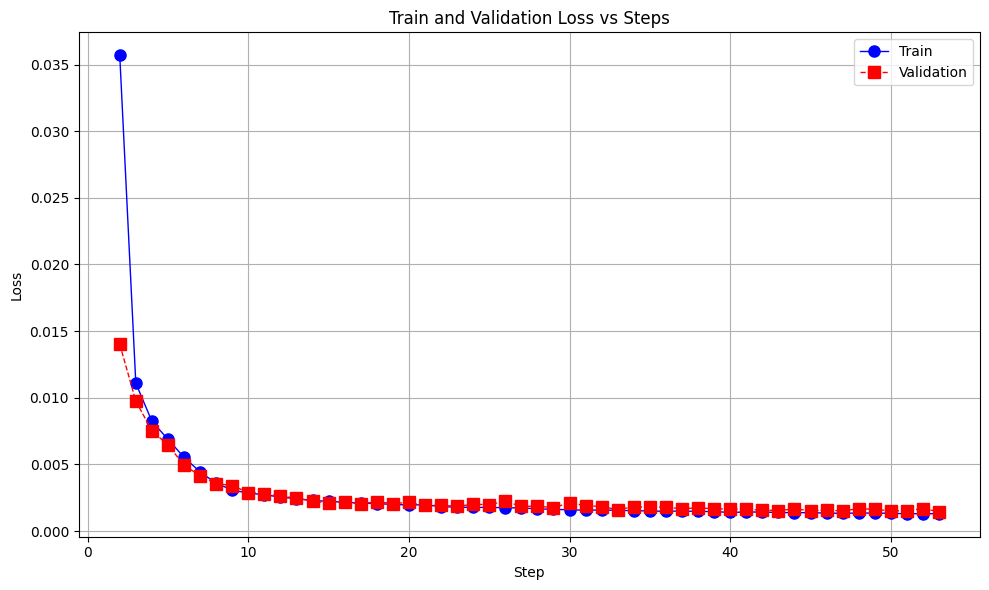}
    \caption{ The training and validation loss values at each step plotted throughout model training. }
    \label{overlayed_loss}
\end{figure}

We assess model performance using Mean Squared Error (MSE), Mean Absolute Percentage Error (MAPE), and Average Confidence metrics. Training and validation losses are logged throughout the epochs, with model weights saved periodically. Figure \ref{overlayed_loss} shows the training and validation loss over time steps.

\subsection{Other Approaches}
The current state-of-the-art for KPR in autonomous racing is the network used by the authors \cite{Dhall2019}. To enable a direct comparison with our UNet-based approach, we replicate their ResNet-based architecture. Instead of training it on their 900 samples with seven keypoints, we use our significantly larger dataset of 20k samples with six keypoints.

We have also implemented the SIFT algorithm \cite{lowe2004sift}. Although there are certain limitations with this method, such as the absence of color calculation and the lack of predictable estimates on predefined keypoints, it can still be used for triangulation. For this, SIFT first identifies unique keypoints in both images and computes descriptors that encapsulate the local image features around each keypoint. These descriptors are then matched using a nearest-neighbor search with FLANN \cite{6809191} and filtered out with log likelihood. The result is a set of corresponding keypoints between the two images.

\subsection{Cone Localization}

After training, the model with the lowest validation loss is loaded to compute the cone location information. We have implemented a disparity-based method for computing the location of the cones. This method works by finding the mean x position of the six keypoints in the stereo image frames and then finding the disparity between them. This can be used to find the depth of the cone with the formula \cite{bradski2008learning}:
\begin{equation}
Z = \frac{f T}{D}
\end{equation}
We get the focal length $f$ from our camera's intrinsic matrix, the baseline $T$ is in the camera specification, and disparity $D$ is found from the keypoints. This depth and the mean x and y positions of the keypoints can be used to generate a point for where the cone is in our coordinate system using:
\begin{equation}
x' = Z
\end{equation}
\begin{equation}
y' = -\frac{(x - c_x) \cdot Z}{f_x}
\end{equation}
\begin{equation}
z' = -\frac{(y - c_y) \cdot Z}{f_y}
\end{equation}
Where $Z$ is depth, $x, y$ are original image coordinates, $c_x, c_y$ are principal points, $f_x, f_y$ are focal lengths, and $x', y', z'$ is the calculated point. This method benefits from using our KPR neural network over other KPR methods. Accurate keypoint correspondences between frames produce a more accurate localization of the cone position. As our KPR method outperforms others in keypoint detection metrics, the localization accuracy is likewise increased.


Beyond position estimation, accurately identifying six keypoints provides a foundation for color estimation, which is essential for path planning to correctly identify track boundaries and driving direction. The six keypoints make a perfect foundation for color estimation, as they clearly identify the stripe (black or white) and base of the cone. Masking can be applied to determine the color of the cone algorithmically. This estimation can be standalone for faster performance or combined with the YOLOv8 class prediction in the downstream filter for higher accuracy and confidence. Note that this color estimation is not possible with SIFT, as it cannot reliably outline the cone's six keypoints.

%% file: sec/3_results.tex
\section{Results}
\label{sec:results}

The final checkpoint was saved and loaded for testing on a set of randomly selected, previously unseen images. The quantitative metrics of our model are evaluated using the MSE metric family and standard deviation of distances between the ground truths and the predicted keypoints. Additionally, we evaluated our model using a combination of precision and recall to output the mean average precision (mAP), adapted to our use case to allow a distance of three pixels within the ground truth to be classified as a correct prediction. The results are shown in Table \ref{table_mse}.

\begin{table}[h]
\caption{Metric Comparison of ResNet and UNet KPR NNs.}
\label{table_mse}
\small 
\setlength{\tabcolsep}{4pt} 
\centering
\begin{tabular}{|c||c|c|c||c||c|}
\hline
\textbf{Model} & \textbf{MSE} & \textbf{Root MSE} & \textbf{Norm ME} & \textbf{Std Dev} & \textbf{mAP} \\
\hline
\textbf{ResNet} & 6.3165 & 2.3458 & 0.0597 & 6.4299 & 0.42 \\ 
\hline
\textbf{UNet} & 3.4172 & 1.8486 & 0.0263 & 3.4550 & 0.83 \\ 
\hline
\end{tabular}
\end{table}




In addition to quantitative results, we evaluate our model under realistic conditions similar to those expected in the competition by employing a recorded ROS bag. The video contains ZED2 stereo camera data obtained from the vehicle during testing. This approach allows us to assess the model’s performance in real-world scenarios that closely mirror its intended application.

\begin{figure}
    \centering
    \includegraphics[width=1\linewidth]{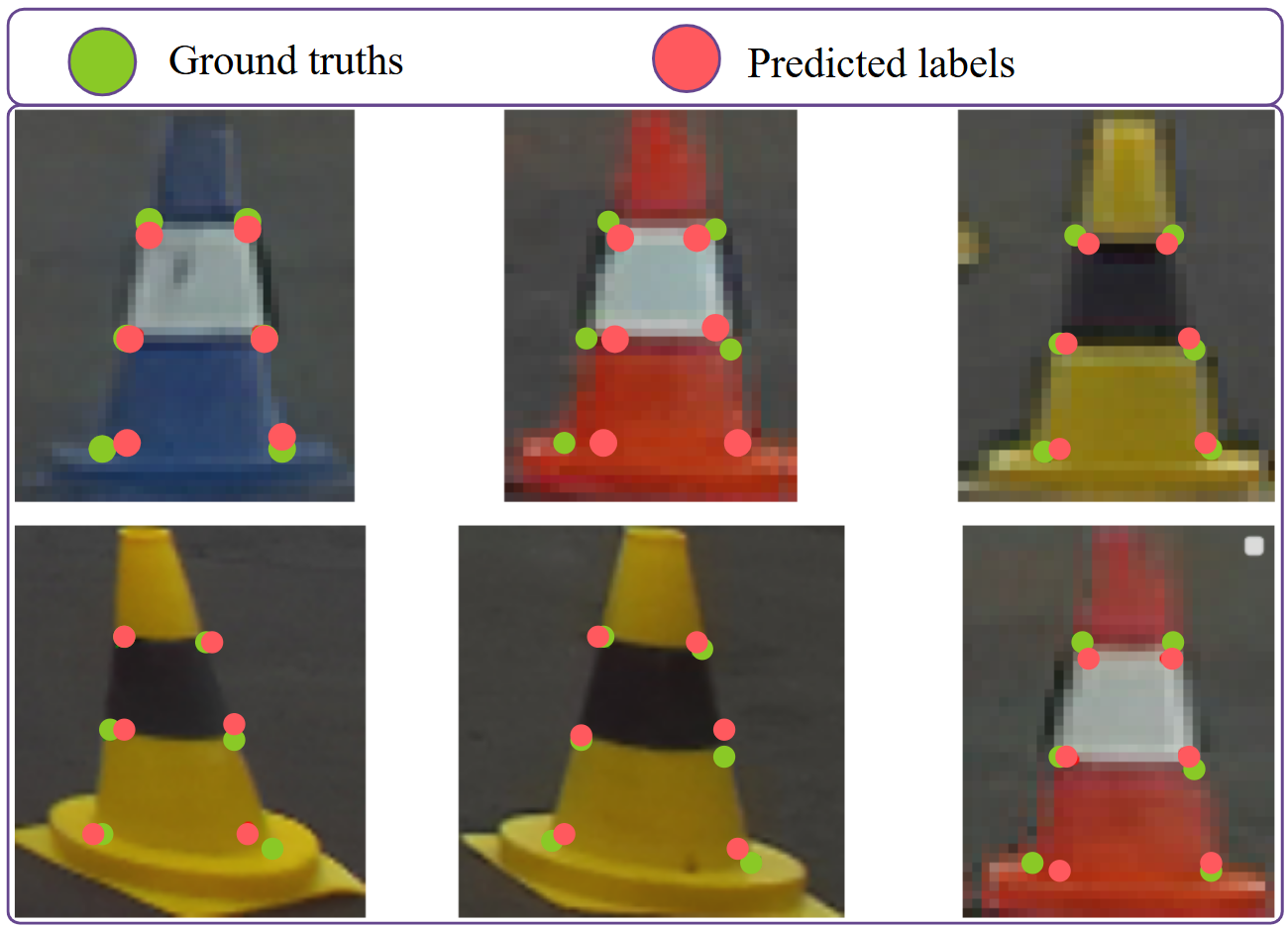}
    \caption{ The mAP measures the distance between true and predicted keypoints indicated by a red line between points.}
    \label{cone_samples}
\end{figure}

The qualitative results are shown in Fig. \ref{cone_samples} with true and predicted keypoints labeled on each image.  

\begin{figure}
    \centering
    \includegraphics[width=1\linewidth]{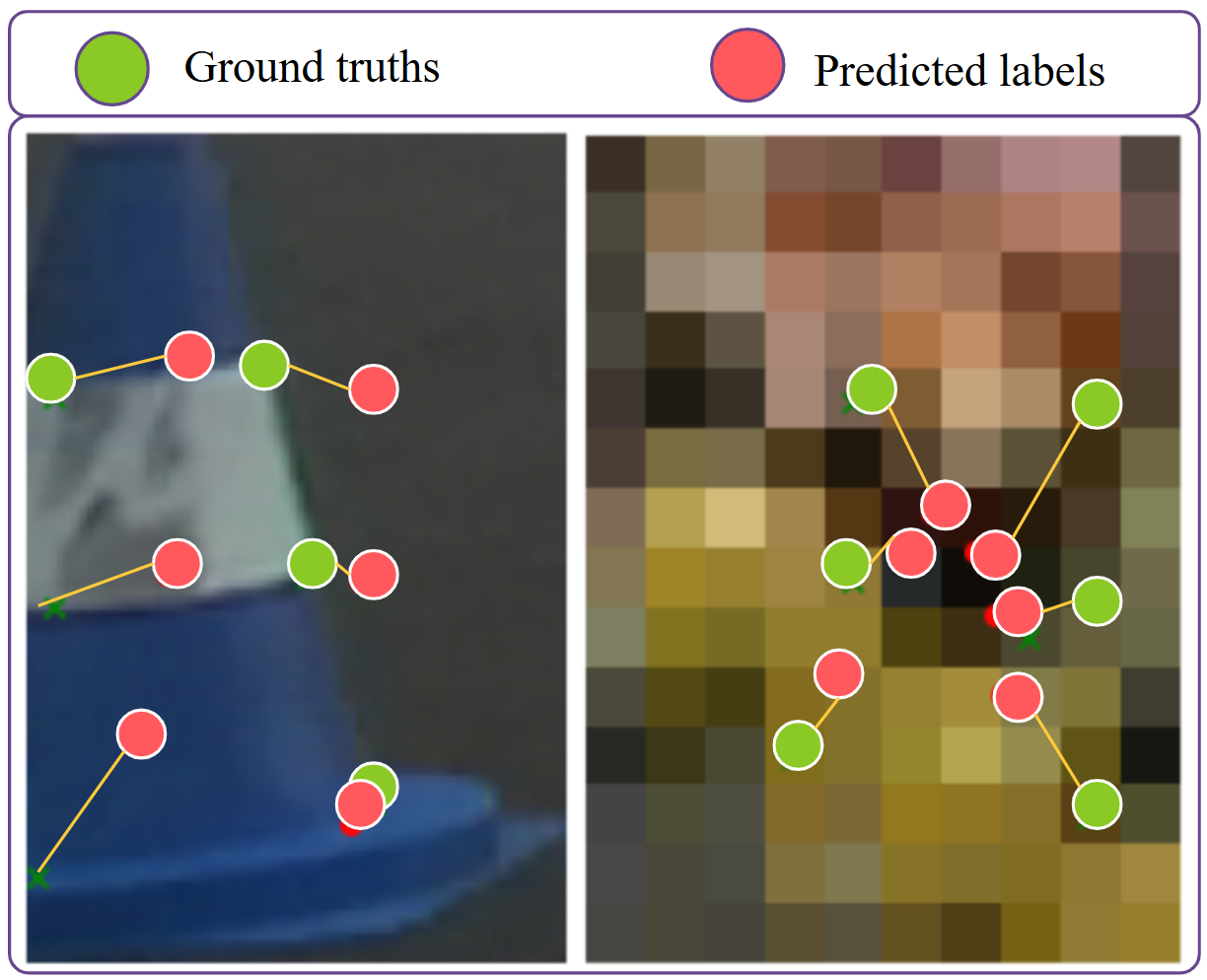}
    \caption{ The predicted keypoints (in red) are inaccurate with respect to the ground truths (in green) due to the cone being partially cropped out by YOLOv8. The yellow line represents the disparity between the labels and inferences. }
    \label{bad_example}
\end{figure}

Even when the KPR NN occasionally fails, it typically does so on challenging cases, such as cones that are either partially out of view or densely clustered within a single image. These challenging cases, as illustrated in Fig. \ref{bad_example}, are relatively rare (approximately 3\% based on our ROS2 bag tests), but they represent the most significant vulnerability in the model. Poor keypoint detection on such samples skews the KPR metrics and possibly propagates into inaccurate depth and racing line calculations further down the autonomous racing pipeline. In the future, we could add such examples to the training data, introduce cropping and translation transformations to our data augmentations, and add an interpretable KPR confidence score to inform the size of the covariance for the Kalman filter downstream.

\subsection{Overall Perception Pipeline}
\begin{figure}
    \centering
    \includegraphics[width=1\linewidth]{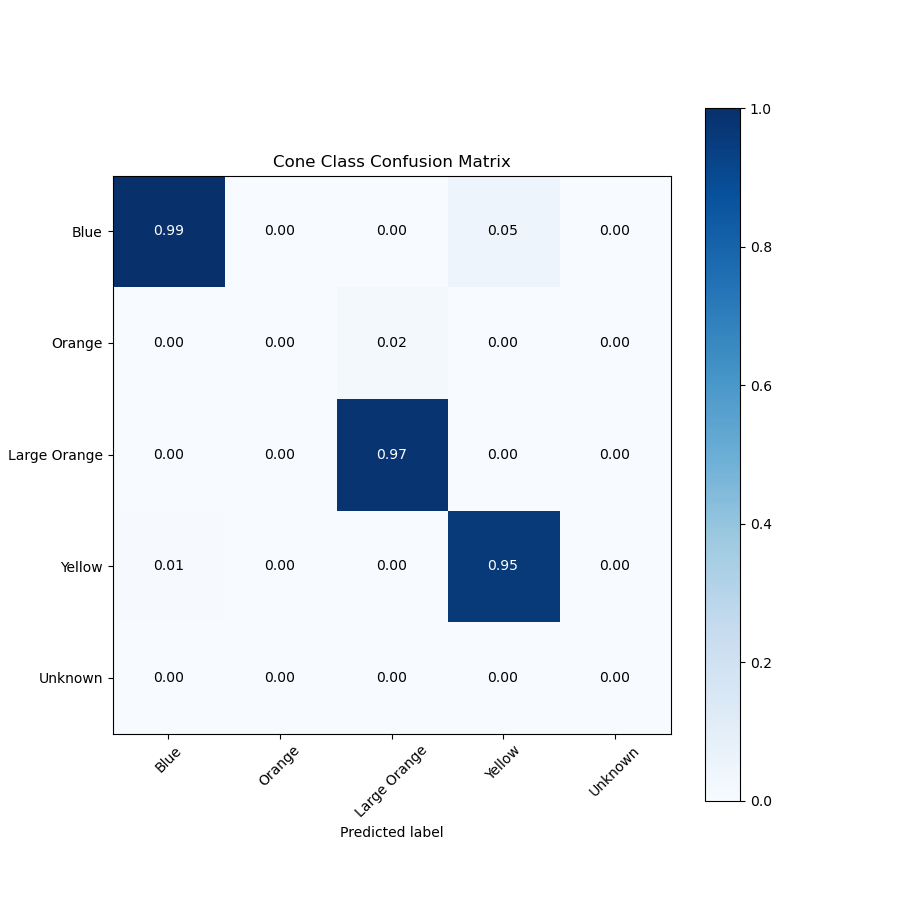}
    \caption{Confusion matrix showing integrated KPR pipeline detected classes against ground truth annotations on simulated data. Off-diagonal non-zero values indicate inaccuracies in the pipeline. }
    \label{eval_confusion_matrix}
\end{figure}

\begin{figure*}[!h]
    \centering
    \includegraphics[width=1\linewidth]{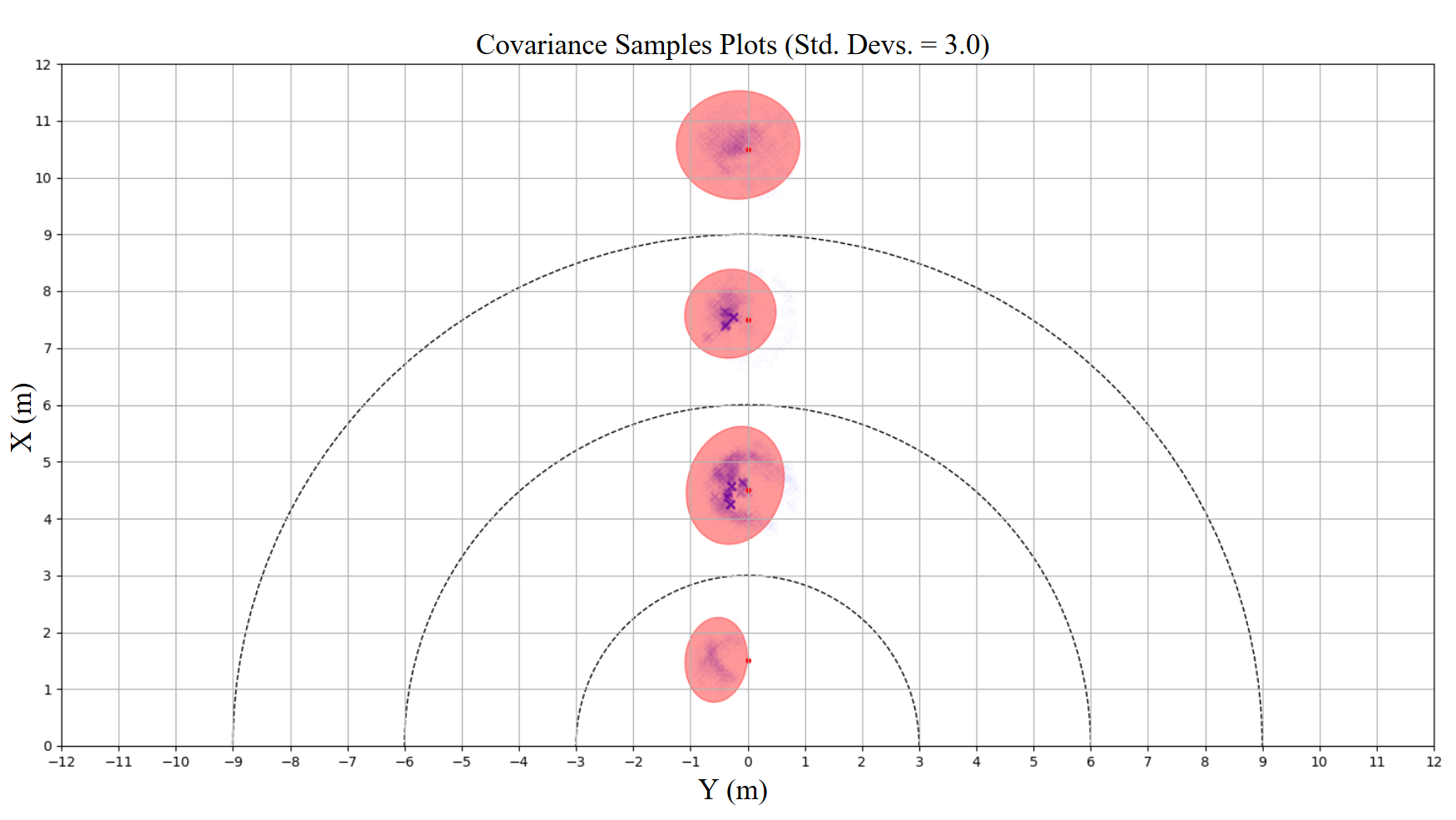}
    \caption{Birds-eye view displaying experimentally derived covariances for perception pipeline at varying distance bins. Blue points represent the error of network predictions, centered around the ground truth position. Red ellipses show covariances for each distance bin, with mean equal to the ground truth position.}
    \label{covariance_samples}
\end{figure*}

To evaluate how the KPR neural network impacts the overall perception pipeline, we run our software on a dedicated simulator with known ground truth poses. We plot the confusion matrix of cone detection accuracy for each cone color in Fig. \ref{eval_confusion_matrix} and a birds-eye view of covariances at varying distances in Fig. \ref{covariance_samples}. The purpose of these metrics is to determine cone detection accuracy for each color. The results shown correspond to our best-performing KPR model integrated into the pipeline with the position estimation method.

\subsection{Real-Time Processing Analysis}

\begin{figure}[H]
    \centering
    \includegraphics[width=1\linewidth]{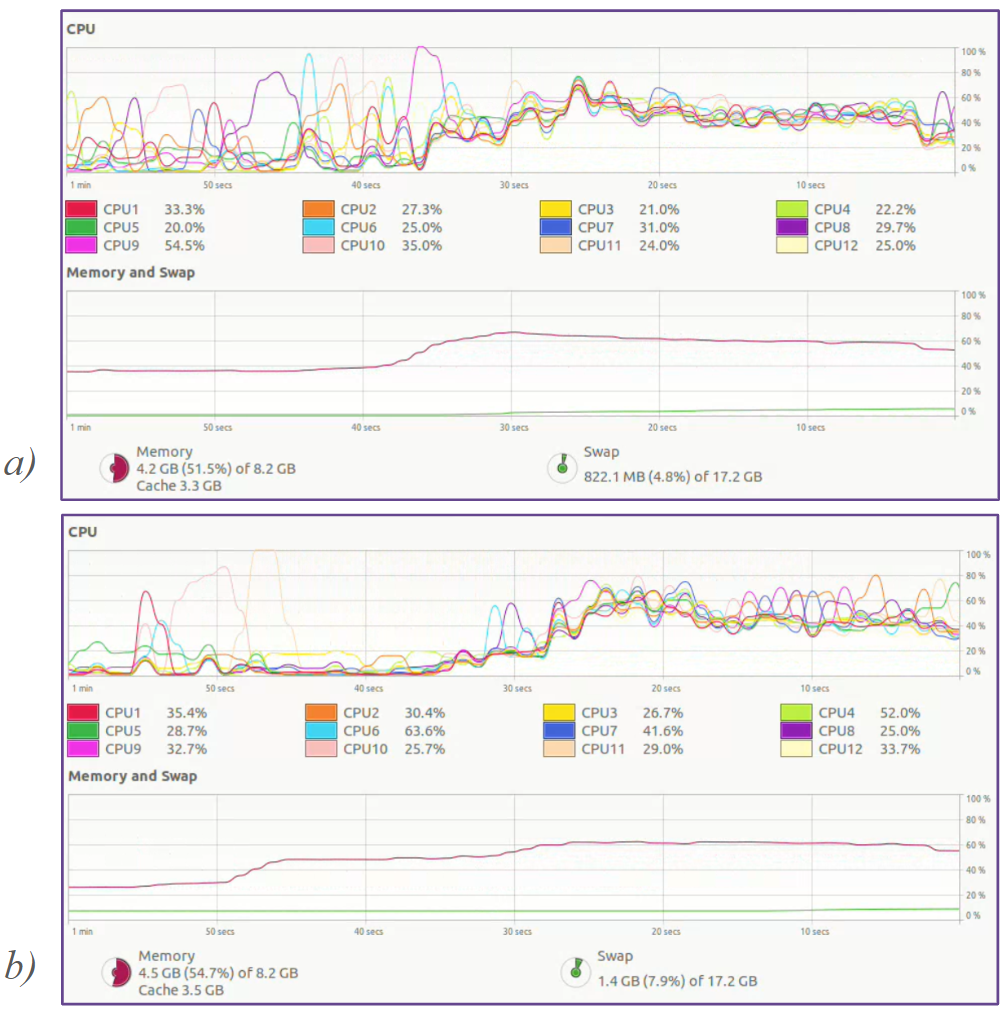}
    \caption{ The CPU and memory usage of the PC inside the car while running a ROS bag. Subfigure (a) shows the plot with no KPR method enabled. Here, the perception pipeline relies on YOLOv8 detection and estimation. The bottom figure (b) runs the pipeline with the KPR neural network enabled. The PC is under heavier computation in this scenario, so the CPU and memory usage become higher. }
    \label{cpu_usage}
\end{figure}

Given that a central contribution of this work is real-time deployment on an autonomous vehicle, we evaluated the processing requirements and runtime feasibility on the in-car PC. Fig. \ref{cpu_usage} illustrates the CPU and memory usage of the in-car PC while processing a ROS2 bag. Subfigure (a) represents the CPU and memory consumption with only YOLOv8 detection and estimation running, while Subfigure (b) shows the additional load when the KPR neural network is enabled. As seen, enabling KPR increases computational demand, placing an increased load on all 12 CPU cores. The memory and swap usage only show a slight increase, with a peak disparity of approximately 7\%, indicating only a marginal impact on overall memory resources. Additionally, peak usage on our GTX 1060 GPU increased by 3\% when using KPR NN from a baseline of 14\%, please see the supplementary material. These lie within our real-time computational requirements for the autonomous racing system.

%% file: sec/4_conclusion.tex
\section{Discussion and Conclusion}
\label{sec:conclusion}
In this paper, we presented a UNet-based neural network for KPR aimed at cone position estimation in autonomous racing applications. By training on a custom-labeled dataset, we demonstrated significant improvements in detection accuracy without a toll on computational efficiency. The model also enables different approaches to depth and color detection for an enhanced overall perception pipeline.

High accuracy in the KPR step of the pipeline is extremely important as the output is used to determine the racing line the car will follow. If a suboptimal path is chosen based on the perception information, the next frames to analyze may in turn have fewer cones, thereby creating a snowball effect. On the other hand, our results show that the addition of a KPR cone estimation method has the potential to amplify system performance in a positive feedback loop. 

Furthermore, we conducted tests on computational efficiency to determine if a machine learning step in the pipeline would hinder the performance. From our test results, we conclude that the marginally higher computational demands are a fair trade-off for the accuracy the KPR provides. This underscores the potential of our KPR model for integration into competitive autonomous racing systems, where its precise and reliable cone detection can amplify overall system performance at subsequent stages of the perception pipeline.

Our approach, with its overarching goal of improving cone localization and depth estimation, presents a promising path forward for real-time, high-speed navigation for autonomous racing. While our method focuses on cone keypoints, the framework can be extended to full 3D scene understanding by regressing depth-aware object maps or semantic voxel grids. Future work may also explore image-text supervision and mitigating occlusion scenarios in a computationally efficient manner.

%% file: main.bib
@String(CVPR= {IEEE Conf. Comput. Vis. Pattern Recog.})

@String(ECCV= {Eur. Conf. Comput. Vis.})

@String(NIPS= {Adv. Neural Inform. Process. Syst.})

@String(CVPR  = {CVPR})

@String(ECCV  = {ECCV})

@String(NIPS  = {NeurIPS})

@article{Contreras2024,
  author = {Contreras, Marcelo and Jain, Aayush and Bhatt, Neel P and Banerjee, Arunava and Hashemi, Ehsan},
  month = {03},
  publisher = {Frontiers Media},
  title = {A survey on 3D object detection in real time for autonomous driving},
  doi = {10.3389/frobt.2024.1212070},
  urldate = {2024-07-22},
  volume = {11},
  year = {2024},
  journal = {Frontiers in robotics and AI}
}

@inbook{Katsamenis2022,
author = {Katsamenis, Iason and Karolou, Eleni and Davradou, Agapi and Protopapadakis, Eftychios and Doulamis, Anastasios and Doulamis, Nikolaos and Kalogeras, Dimitris},
year = {2022},
month = {09},
pages = {382-391},
title = {TraCon: A Novel Dataset for Real-Time Traffic Cones Detection Using Deep Learning},
isbn = {978-3-031-17600-5},
doi = {10.1007/978-3-031-17601-2_37}
}

@INPROCEEDINGS{Dhall2019,
  author={Dhall, Ankit and Dai, Dengxin and Van Gool, Luc},
  booktitle={2019 IEEE Intelligent Vehicles Symposium (IV)}, 
  title={Real-time 3D Traffic Cone Detection for Autonomous Driving}, 
  year={2019},
  volume={},
  number={},
  pages={494-501},
  doi={10.1109/IVS.2019.8814089}
}

@InProceedings{Martinez2022,
author="Albaranez Martinez, Javier
and Llopis-Ibor, Laura
and Hernandez-Garcia, Sergio
and Pineda de Luelmo, Susana
and Hernandez-Ferrandiz, Daniel",
editor="Pinho, Armando J.
and Georgieva, Petia
and Teixeira, Luis F.
and Sanchez, Joan Andreu",
title="A Case of Study on Traffic Cone Detection for Autonomous Racing on a Jetson Platform",
booktitle="Pattern Recognition and Image Analysis",
year="2022",
publisher="Springer International Publishing",
address="Cham",
pages="629--641",
isbn="978-3-031-04881-4"
}

@InProceedings{Sun2018,
author="Sun, Xiao
and Xiao, Bin
and Wei, Fangyin
and Liang, Shuang
and Wei, Yichen",
editor="Ferrari, Vittorio
and Hebert, Martial
and Sminchisescu, Cristian
and Weiss, Yair",
title="Integral Human Pose Regression",
booktitle="Computer Vision -- ECCV 2018",
year="2018",
publisher="Springer International Publishing",
address="Cham",
pages="536--553",
isbn="978-3-030-01231-1"
}

@InProceedings{Zhang2019,
author = {Zhang, Feng and Zhu, Xiatian and Ye, Mao},
title = {Fast Human Pose Estimation},
booktitle = {Proceedings of the IEEE/CVF Conference on Computer Vision and Pattern Recognition (CVPR)},
month = {June},
year = {2019}
}

@ARTICLE{Liang2023,
  author={Liang, Xu and Fan, Dandan and Yang, Jinyang and Jia, Wei and Lu, Guangming and Zhang, David},
  journal={IEEE Journal of Selected Topics in Signal Processing}, 
  title={PKLNet: Keypoint Localization Neural Network for Touchless Palmprint Recognition Based on Edge-Aware Regression}, 
  year={2023},
  volume={17},
  number={3},
  pages={662-676},
  doi={10.1109/JSTSP.2023.3241540}}

@inproceedings{Yu2024,
author = {Yu, Dongyang and Xie, Yunshi and An, Wangpeng and Li, Zhang and Yao, Yufeng},
title = {Joint Coordinate Regression and Association For Multi-Person Pose Estimation, A Pure Neural Network Approach},
year = {2024},
isbn = {9798400702051},
publisher = {Association for Computing Machinery},
address = {New York, NY, USA},
url = {https://doi.org/10.1145/3595916.3626382},
doi = {10.1145/3595916.3626382},
booktitle = {Proceedings of the 5th ACM International Conference on Multimedia in Asia},
articleno = {13},
numpages = {8},
location = {Tainan, Taiwan},
series = {MMAsia '23}
}

@ARTICLE{6809191,
  author={Muja, Marius and Lowe, David G.},
  journal={IEEE Transactions on Pattern Analysis and Machine Intelligence}, 
  title={Scalable Nearest Neighbor Algorithms for High Dimensional Data}, 
  year={2014},
  volume={36},
  number={11},
  pages={2227-2240},
  doi={10.1109/TPAMI.2014.2321376}
}

@Article{s24227347,
AUTHOR = {Li, Mingjing and Liu, Xinyang and Chen, Shuang and Yang, Le and Du, Qingyu and Han, Ziqing and Wang, Junshuai},
TITLE = {MST-YOLO: Small Object Detection Model for Autonomous Driving},
JOURNAL = {Sensors},
VOLUME = {24},
YEAR = {2024},
NUMBER = {22},
ARTICLE-NUMBER = {7347},
URL = {https://www.mdpi.com/1424-8220/24/22/7347},
PubMedID = {39599123},
ISSN = {1424-8220},
DOI = {10.3390/s24227347}
}

@article{Wang2023, 
author={Wang, Xiuling and Kong, Lingkun and Zhang, Zhiguo and Wang, Haixia and Lu, Xiao}, 
title={Keypoint regression strategy and angle loss based YOLO for object detection}, 
year={2023}, 
month={Nov},
volume={13}, 
doi={https://doi.org/10.1038/s41598-023-47398-w}, 
number={1}, 
journal={Scientific reports}, 
publisher={Nature Portfolio}, 
}

@ARTICLE{Duan2024,
  author={Duan, Kaiwen and Bai, Song and Xie, Lingxi and Qi, Honggang and Huang, Qingming and Tian, Qi},
  journal={IEEE Transactions on Pattern Analysis and Machine Intelligence}, 
  title={CenterNet++ for Object Detection}, 
  year={2024},
  volume={46},
  number={5},
  pages={3509-3521},
  doi={10.1109/TPAMI.2023.3342120}}

@INPROCEEDINGS{Lu2022,
  author={Lu, Dongchen and Li, Dongmei and Li, Yali and Wang, Shengjin},
  booktitle={2022 IEEE/CVF Conference on Computer Vision and Pattern Recognition (CVPR)}, 
  title={OSKDet: Orientation-sensitive Keypoint Localization for Rotated Object Detection}, 
  year={2022},
  volume={},
  number={},
  pages={1172-1182},
  doi={10.1109/CVPR52688.2022.00125}}

@InProceedings{Nguyen2014,
author="Nguyen, Thao
and Park, Eun-Ae
and Han, Jiho
and Park, Dong-Chul
and Min, Soo-Young",
editor="Pan, Jeng-Shyang
and Kr{\"o}mer, Pavel
and Sn{\'a}{\v{s}}el, V{\'a}clav",
title="Object Detection Using Scale Invariant Feature Transform",
booktitle="Genetic and Evolutionary Computing",
year="2014",
publisher="Springer International Publishing",
address="Cham",
pages="65--72",
isbn="978-3-319-01796-9"
}

@INPROCEEDINGS{Nandhini2023,
  author={Nandhini, T J and Thinakaran, K},
  booktitle={2023 Fifth International Conference on Electrical, Computer and Communication Technologies (ICECCT)}, 
  title={SIFT algorithm-based Object detection and tracking in the video image}, 
  year={2023},
  volume={},
  number={},
  pages={1-4},
  doi={10.1109/ICECCT56650.2023.10179720}}

@InProceedings{Rosten2006,
author="Rosten, Edward
and Drummond, Tom",
editor="Leonardis, Ale{\v{s}}
and Bischof, Horst
and Pinz, Axel",
title="Machine Learning for High-Speed Corner Detection",
booktitle="Computer Vision -- ECCV 2006",
year="2006",
publisher="Springer Berlin Heidelberg",
address="Berlin, Heidelberg",
pages="430--443",
isbn="978-3-540-33833-8"
}

@article{Rosten2010,
author = {Rosten, Edward and Porter, Reid and Drummond, Tom},
year = {2010},
month = {01},
pages = {105-19},
title = {FASTER and better: A Machine Learning Approach to Corner Detection},
volume = {32},
journal = {IEEE transactions on pattern analysis and machine intelligence},
doi = {10.1109/TPAMI.2008.275}
}

@misc{loshchilov2019decoupledweightdecayregularization,
      title={Decoupled Weight Decay Regularization}, 
      author={Ilya Loshchilov and Frank Hutter},
      year={2019},
      eprint={1711.05101},
      archivePrefix={arXiv},
      primaryClass={cs.LG},
      url={https://arxiv.org/abs/1711.05101}, 
}

@article{mohammed2020perception,
  title={The perception system of intelligent ground vehicles in all weather conditions: A systematic literature review},
  author={Mohammed, Abdul Sajeed and Amamou, Ali and Ayevide, Follivi Kloutse and Kelouwani, Sousso and Agbossou, Kodjo and Zioui, Nadjet},
  journal={Sensors},
  volume={20},
  number={22},
  pages={6532},
  year={2020},
  publisher={MDPI}
}

@article{fursa2021worsening,
  title={Worsening perception: Real-time degradation of autonomous vehicle perception performance for simulation of adverse weather conditions},
  author={Fursa, Ivan and Fandi, Elias and Musat, Valentina and Culley, Jacob and Gil, Enric and Teeti, Izzeddin and Bilous, Louise and Sluis, Isaac Vander and Rast, Alexander and Bradley, Andrew},
  journal={arXiv preprint arXiv:2103.02760},
  year={2021}
}

@inproceedings{maji2022yolo,
  title={Yolo-pose: Enhancing yolo for multi person pose estimation using object keypoint similarity loss},
  author={Maji, Debapriya and Nagori, Soyeb and Mathew, Manu and Poddar, Deepak},
  booktitle={Proceedings of the IEEE/CVF Conference on Computer Vision and Pattern Recognition},
  pages={2637--2646},
  year={2022}
}

@inproceedings{lin2014microsoft,
  title={Microsoft coco: Common objects in context},
  author={Lin, Tsung-Yi and Maire, Michael and Belongie, Serge and Hays, James and Perona, Pietro and Ramanan, Deva and Doll{\'a}r, Piotr and Zitnick, C Lawrence},
  booktitle={Computer Vision--ECCV 2014: 13th European Conference, Zurich, Switzerland, September 6-12, 2014, Proceedings, Part V 13},
  pages={740--755},
  year={2014},
  organization={Springer}
}

@inproceedings{bellavia2019evaluation,
  title={Which is Which? Evaluation of local descriptors for image matching in real-world scenarios},
  author={Bellavia, Fabio and Colombo, Carlo},
  booktitle={Computer Analysis of Images and Patterns: 18th International Conference, CAIP 2019, Salerno, Italy, September 3--5, 2019, Proceedings, Part I 18},
  pages={299--310},
  year={2019},
  organization={Springer}
}

@article{bay2008speeded,
  title={Speeded-up robust features (SURF)},
  author={Bay, Herbert and Ess, Andreas and Tuytelaars, Tinne and Van Gool, Luc},
  journal={Computer vision and image understanding},
  volume={110},
  number={3},
  pages={346--359},
  year={2008},
  publisher={Elsevier}
}

@inproceedings{sarlin2020superglue,
  title={Superglue: Learning feature matching with graph neural networks},
  author={Sarlin, Paul-Edouard and DeTone, Daniel and Malisiewicz, Tomasz and Rabinovich, Andrew},
  booktitle={Proceedings of the IEEE/CVF conference on computer vision and pattern recognition},
  pages={4938--4947},
  year={2020}
}

@inproceedings{he2016deep,
  title={Deep residual learning for image recognition},
  author={He, Kaiming and Zhang, Xiangyu and Ren, Shaoqing and Sun, Jian},
  booktitle={Proceedings of the IEEE conference on computer vision and pattern recognition},
  pages={770--778},
  year={2016}
}

@inproceedings{o2020deep,
  title={Deep learning vs. traditional computer vision},
  author={O’Mahony, Niall and Campbell, Sean and Carvalho, Anderson and Harapanahalli, Suman and Hernandez, Gustavo Velasco and Krpalkova, Lenka and Riordan, Daniel and Walsh, Joseph},
  booktitle={Advances in Computer Vision: Proceedings of the 2019 Computer Vision Conference (CVC), Volume 1 1},
  pages={128--144},
  year={2020},
  organization={Springer}
}

@article{bradski2008learning,
  title={Learning OpenCV: Computer vision with the OpenCV library},
  author={Bradski, Gary},
  journal={O'REILLY google schola},
  volume={2},
  pages={334--352},
  year={2008}
}

@article{lowe2004sift,
  title={Sift-the scale invariant feature transform},
  author={Lowe, G},
  journal={Int. J},
  volume={2},
  number={91-110},
  pages={2},
  year={2004}
}

@article{fsoco_2022,
  title={FSOCO: The Formula Student Objects in Context Dataset},
  author={V{\"o}disch, Niclas and Dodel, David and Sch{\"o}tz, Michael},
  journal={SAE International Journal of Connected and Automated Vehicles},
  volume={5},
  number={12-05-01-0003},
  year={2022}
}

@inproceedings{mildenhall2020nerf,
  title={NeRF: Representing Scenes as Neural Radiance Fields for View Synthesis},
  author={Mildenhall, Ben and Srinivasan, Pratul P and Tancik, Matthew and Barron, Jonathan T and Ramamoorthi, Ravi and Ng, Ren},
  booktitle={ECCV},
  pages={405--421},
  year={2020},
  publisher={Springer}
}

@inproceedings{eigen2014depth,
  title={Depth Map Prediction from a Single Image using a Multi-Scale Deep Network},
  author={Eigen, David and Puhrsch, Christian and Fergus, Rob},
  booktitle={NIPS},
  pages={2366--2374},
  year={2014}
}
